\documentclass[letterpaper, 10 pt, conference]{ieeeconf}  % Comment this line out if you need a4paper

\IEEEoverridecommandlockouts 

\usepackage{graphicx}
\usepackage{glossaries}
\usepackage{amsfonts}
\usepackage{caption}
\usepackage{subcaption}
\usepackage{tabularx}
\usepackage{algorithm2e}
\usepackage{tikz}   % tikz
\RestyleAlgo{ruled}

\usepackage{amsmath}

\newcommand*\numcircled[1]{\tikz[baseline=(char.base)]{
            \node[shape=circle,draw,inner sep=0.75pt] (char) {#1};}} 

\begin{document}

\title{\LARGE \bf Hybrid Tendon and Ball Chain Continuum Robots for Enhanced Dexterity in Medical Interventions}

\author{Giovanni~Pittiglio$^1$,  Margherita~Mencattelli$^1$, Abdulhamit~Donder$^1$, Yash~Chitalia$^2$, and Pierre~E.~Dupont$^1$
\thanks{$^1$Department of Cardiovascular Surgery, Boston Children’s Hospital, Harvard Medical School, Boston, MA 02115, USA. Email:
        {\tt\small \{giovanni.pittiglio, margherita.mencattelli, abdulhamit.donder, pierre.dupont\}@childrens.harvard.edu}
        \newline $^2$Department of Mechanical Engineering, University of Louisville, Louisville, KY 40292, USA.
 {\tt\small yash.chitalia@louisville.edu}
        \newline This work was supported by the National Institutes of Health under grant R01HL124020.}%
}

\maketitle

\begin{abstract}
A hybrid continuum robot design is introduced that combines a proximal tendon-actuated section with a distal telescoping section comprised of permanent-magnet spheres actuated using an external magnet. While, individually, each section can approach a point in its workspace from one or at most several orientations, the two-section combination possesses a dexterous workspace. The paper describes kinematic modeling of the hybrid design and provides a description of the dexterous workspace. We present experimental validation which shows that a simplified kinematic model produces tip position mean and maximum errors of 3\% and 7\% of total robot length, respectively.

\end{abstract}


\IEEEpeerreviewmaketitle
\section{Introduction}
Continuum robots have attracted considerable attention for applications in minimally invasive diagnostics and therapeutics over the past decade \cite{Dupont2022}. The primary reason is their ability to navigate narrow and tortuous anatomical passageways, while safely interacting with the anatomy. 

In designing such robots, an important goal is to create a robot with a workspace appropriate for the clinical task. A significant limitation of many continuum designs is that these robots lack a dexterous workspace. While clinical applications may necessitate approaching a target with a specific angle of approach, continuum designs are often limited in this regard. Furthermore, while multiple bending sections can be concatenated to provide more degrees of freedom, the orientations by which a point in the workspace can be approached are often highly constrained. 

\begin{figure}[t]
    \centering
    \includegraphics[width = \columnwidth]{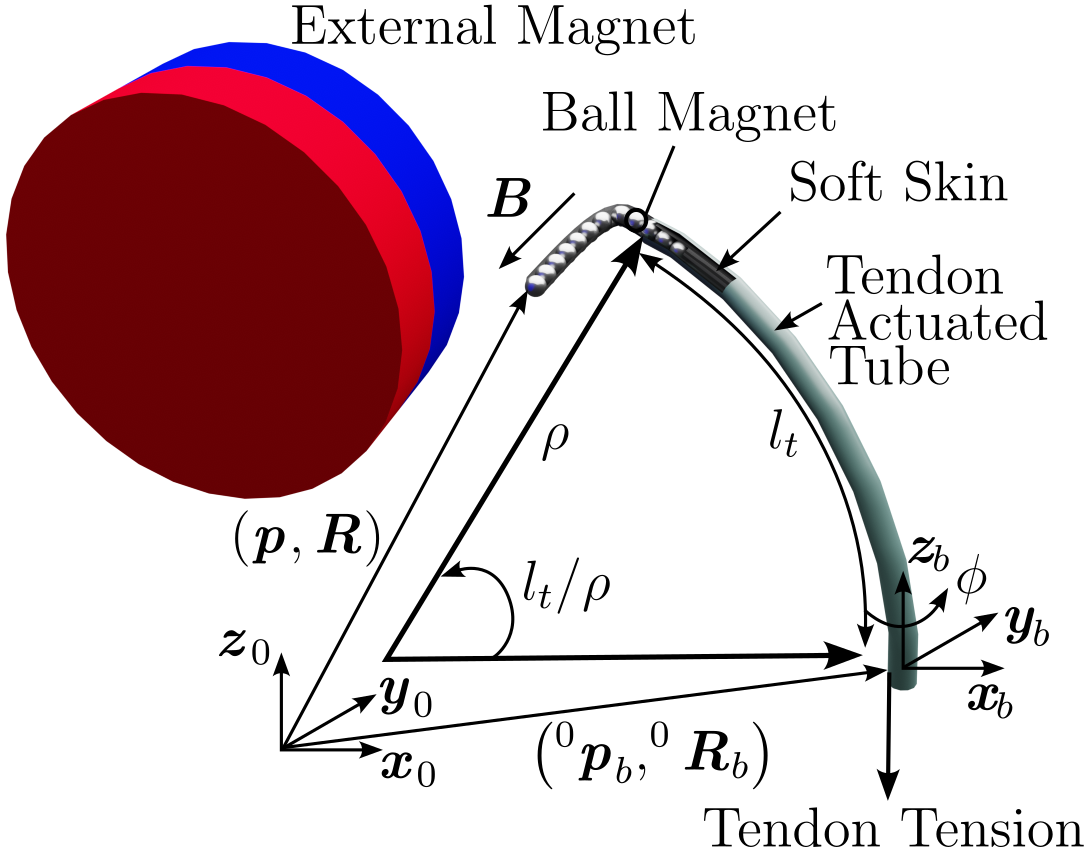}
    \caption{Hybrid robot comprised of proximal tendon-actuated section and distal telescoping magnetic ball-chain section.}
    \label{fig:hybrid}
\end{figure}

To overcome this limitation, this paper investigates a hybrid design that combines the advantages of tendon actuation \cite{Rucker2011} and magnetic ball chain robots \cite{Pittiglio2023a, ODonoghue2013} as shown in Fig. \ref{fig:hybrid}. In this hybrid design, a proximal tendon-actuated section positions the robot with respect to the goal tip location while a distal ball chain section orients the robot tip via an externally-produced magnetic field.

This hybrid design is perhaps the only two section continuum robot possessing a true dexterous workspace - a continuous set of tip positions that can be approached from an arbitrary direction. %{\color{red} This results fundamental in applications, where both positioning and orientating the robot's tip is important: biopsy, excising lesions, and cardiac ablation.}

The remainder of the paper is arranged as follows. The next two sections describe how kinematic models of varying complexity can be constructed for the tendon- and magnetically-actuated design. The subsequent section uses an approximate kinematic model to characterize the dexterous workspace. Section V contains an experimental validation of the kinematic model and workspace. Conclusions appear in the final section.

% \begin{figure}[t]
%     \centering
%     \includegraphics[width = 0.75\columnwidth]{figures/bladder.png}
%     \caption{Adding a retractable tendon-actuated tube to the ball chain steerable tip enables the robot to approach a tip position, $p$, from multiple tangent directions.}
%     \label{fig:main}
% \end{figure}

\section{Kinematic Modeling}
\label{sec:static_mod}
A complete mechanics-based model of the hybrid design can be derived by combining the tendon-actuated model of  \cite{Rucker2011} with the magnetic energy ball-chain model of \cite{Pittiglio2023a, ODonoghue2013}. In this approach, the magnetic balls retracted inside the tendon-actuated tube exert ``external'' point loads on the tube while the extended portion of the ball chain exerts a tip force on the tube. The magnitudes of these forces can be computed using the formulation of \cite{Pittiglio2023a}, but depend on the deflected tube shape. An iterative  approach can be used to alternately solve the tube and ball chain models until convergence is reached. 

If the forces from the magnetic ball chain do not appreciably deflect the tendon-actuated tube, the kinematic modeling equations and solution procedure can be substantially simplified and solved in a single iteration. In this case, the tendon-actuated tube shape can be computed first. The position constraints from the tube shape can then be imposed on the ball chain to solve for the shape of its extended portion. When this simplification can be applied, it is advantageous since the model becomes less complex and the computation time faster. Because of these advantages, we investigate the simplified model in this paper. The following subsections summarize the tendon-actuated tube model and the magnetic energy ball chain model.

\subsection{Tendon-actuated tube}
\label{sub:mech_mod}
% \begin{figure}
%     \centering
%     \includegraphics{figures/mech_model.png}
%     \caption{Description of the load on the tendon actuated tube.}
%     \label{fig:mech_model}
% \end{figure}
The centerline of the tube can be described, as a  function of arc-length, $s \in \left[0, l_t\right]$, with its position $\pmb p_c(s) \in \mathbb{R}^3$ and orientation $\pmb R_c(s) \in SO(3)$,
\begin{subequations}
\label{eq:tube_cent}
    \begin{align}
    \dot{\pmb p}_c(s) &= \pmb R_c(s) \pmb v(s) \\
    \dot{\pmb R}_c(s) &= \pmb R_c(s) \pmb u_\times(s)
   \end{align}
\end{subequations}
where $\pmb v(s) \in \mathbb{R}^3$ and $\pmb u(s) \in \mathbb{R}^3$ are the respective linear and angular strain; the ``skew" operator is defined as $\pmb{u}_\times = \left(\pmb{u} \times \pmb e_1  \ \pmb{u} \times \pmb e_2 \ \pmb{u} \times \pmb e_3  \right)$ with $\pmb e_i\in \mathbb{R}^{3}$ $i$th element of the canonical basis of $\mathbb{R}^3$, $\times$ indicating the cross product. Here, the dot notation is used to indicate derivatives with respect to arc-length (i.e. $\dot{\pmb p}_c(s) = \frac{d{\pmb p}_c(s)}{ds}$). Following the Cosserat rod modeling \cite{Rucker2011}, we describe the internal forces $\pmb n$ and moments $\pmb m$ with respect to the applied force $\pmb f$ and torque $\pmb \tau$ as
\begin{equation}
\label{eq:int_forces}
\left(
\begin{array}{c}
    \dot{\pmb n}  \\
     \dot{\pmb m} 
\end{array}
\right)
= 
\left(
\begin{array}{cc}
    \pmb u_\times & \pmb 0 \\
    \pmb v_\times & \pmb{u}_\times
\end{array}
\right)
\left(
\begin{array}{c}
    \pmb n  \\
    \pmb m
\end{array}
\right)
+ 
\left(
\begin{array}{c}
    \pmb f  \\
    \pmb \tau
\end{array}
\right)
\end{equation}
Under assumption of linear elasticity, we have
\begin{equation}
    \left(
\begin{array}{c}
    \pmb n  \\
    \pmb m
\end{array}
\right) = K 
\left(
\begin{array}{c}
    \pmb u - \pmb{u}_0 \\
    \pmb v - \pmb{v}_0
\end{array}
\right)
\end{equation}
with $K = diag(GA, GA, EA, EI_{xx}, EI_{yy}, GJ_{zz})$, where $G$ is the shear modulus, $E$ modulus of elasticity, $A$ cross-sectional area, $I_{xx}$, $I_{yy}$  second moment of area around the $\pmb x$ and $\pmb y$ axes, and $J_{zz}$ the polar moment of area; $\pmb{u}_0$ and $\pmb{v}_0$ are the initial angular and linear strain respectively.

The tube is subject to two main external inputs: distributed force related to the force applied by the tendon at the tip  $\pmb f_t$; and gravitational force $\pmb f_g = m \pmb g$, $m$ mass density and $\pmb g$ gravitational acceleration.

Following \cite{Rucker2011}, the path of the tendon on the surface of the tube along the $\pmb x$ axis is
\begin{equation}
    \pmb p_t(s) = \pmb p_c(s) + \frac{d_t}{2} \pmb R_c(s) \pmb{e}_1
\end{equation}
and the relative force and torque are
\begin{subequations}
\begin{align}
     \pmb f_t &= -\lambda \frac{\dot{\pmb{p}}^2_{t_\times}}{\|\pmb{p}_{t}\|^3}\ddot{\pmb p}_t \\
     \pmb{\tau}_t &= \pmb{p}_t \times \pmb f_t
\end{align}
\end{subequations}
with $\lambda$ tension applied to the tendon and $d_t$ is the diameter of the tube.

The load at the tip is implemented as a boundary condition,
\begin{subequations}
\label{eq:mech_bound}
\begin{align}
    \pmb F(l_t) &= -\lambda \frac{\dot{\pmb p}_c(l_t)}{\|\dot{\pmb p}_c(l_t)\|} \\
    \pmb T(l_t) &= (\pmb p_c(l_t) - \pmb p_t(l_t)) \times \pmb F(l_t).
\end{align}
\end{subequations}

\subsection{Magnetic ball chain}
\label{sub:mag_mod}
The equilibrium configuration of the magnetic ball chain can be solved by minimizing the potential energy of the system. Following \cite{Pittiglio2023a, Pittiglio2023b}, the total potential energy of the balls can be expressed as a combination of magnetic and gravitational components as 
\begin{subequations}
\label{eq:energies}
    \begin{align}
       U_{e} &= -\sum_{i = 1}^n {} \pmb{\mu}_i \cdot {} \pmb{B}(\pmb{p}_i - \pmb{p}_e, \pmb{\mu}_e ) \label{seq:ext_mag} \\
        U_b &= -\sum_{i = 1}^n \sum_{j = i + 1}^n {} \pmb{\mu}_i \cdot \pmb{B}(\pmb{p}_i -\pmb{p}_j, \pmb{\mu}_j ) \label{seq:int_mag} \\ 
        U_g &= \sum_{i = 1}^n m_i \pmb{g}^T \pmb{p}_i, \label{seq:grav}
    \end{align}
\end{subequations}
In (\ref{seq:ext_mag}), we express the interaction with the external magnet in position $\pmb p_e$ and magnetic dipole $\pmb \mu_e$, (\ref{seq:int_mag}) considers the intermagnet energy for the magnets in the chain, and (\ref{seq:grav}) is the gravitational component. The magnetic field is computed using the dipole model
$$\pmb{B}(\pmb{r}, \pmb{\mu}) = \frac{ \mu_0}{4 \pi |\pmb{r}|^3} \left(3 \hat{\pmb{r}} \hat{\pmb{r}}^T - I \right) \pmb{\mu}.$$

\begin{figure}[t]
    \centering
    \includegraphics[width = \columnwidth]{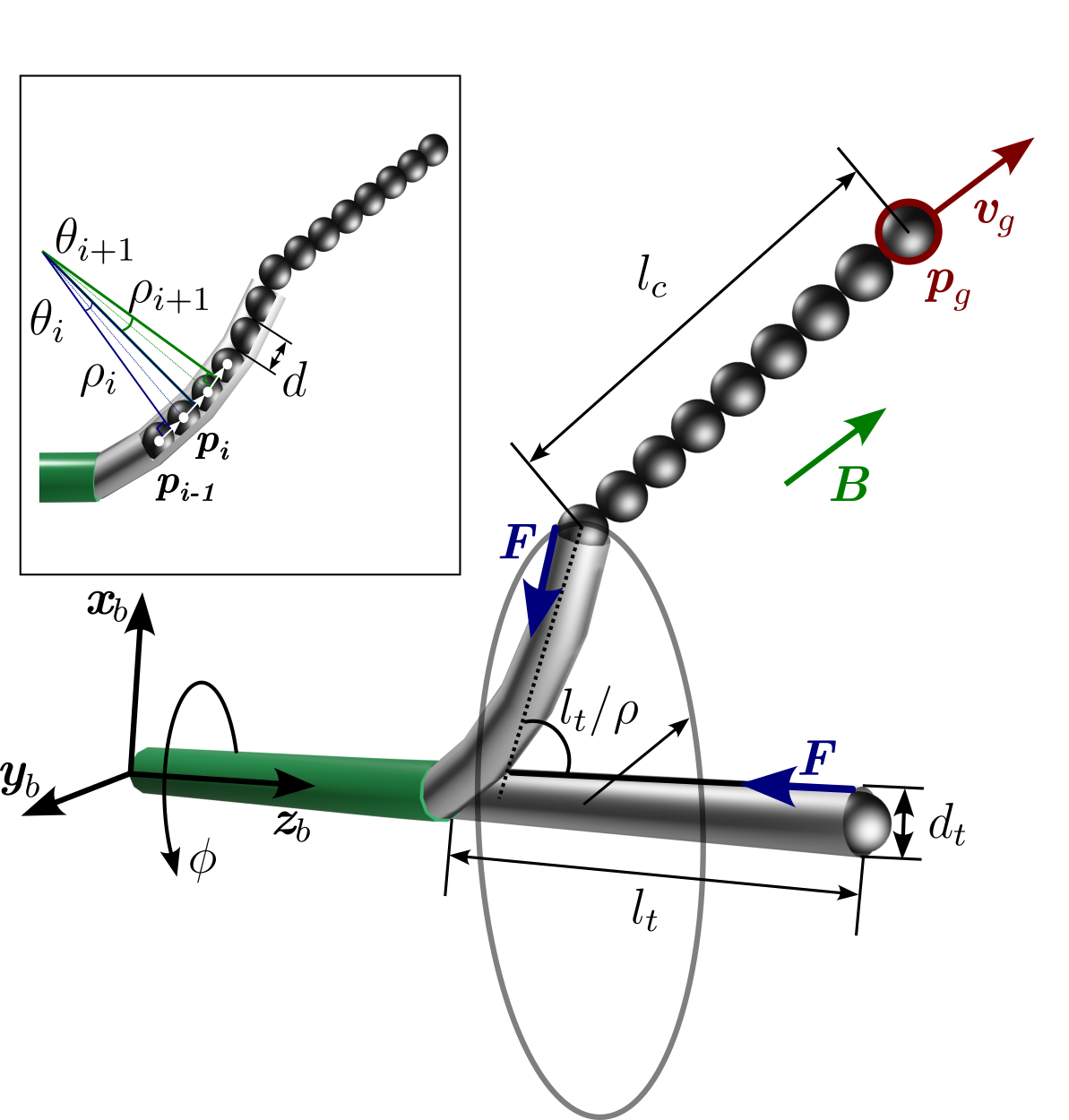}
    \caption{Hybrid model parameters. Proximal section, $l_t$, is assumed constant curvature with radius of curvature $\rho$ under tension $F$. Extended ball chain, $l_c$, is assumed linear in field direction $\pmb B$.}
    \label{fig:inv_kin}
\end{figure}

By combining the terms in (\ref{eq:energies}), we obtain the total energy $U = U_e + U_b  + U_g$.
In \cite{Pittiglio2023a},  we have considered the discrete nature of the chain and discretized the radius of curvature $\rho_i$ and bending angle $\theta_i$ (see Fig. \ref{fig:inv_kin}) between consecutive balls, to consider elastic terms in the sleeve containing the chain. In contrast to our prior work, we solve for the minimum energy configuration while constraining the position of the $n_i$ balls inside the tube using Matlab function {\it fmincon}, since the tube can be considered infinitely stiff, in this case. These constraints are given by 
\begin{equation}
\label{eq:constraints_simp}
        \pmb{p}_i = \pmb{p}_{i_0} \ \forall \ i=1,2,\dots,n_i
\end{equation}
$\pmb{p}_{i_0}$ is derived from the the Cosserat equilibrium.

As in our previous work \cite{Pittiglio2023b}, we constrain each ball to be in continuous contact with the previous one and the dipole intensity of the $n$ balls to be constant
\begin{subequations}
    \begin{align}
        \|\pmb{p}_i - \pmb{p}_{i-1}\| &= d \ \forall \ i=2,3,\dots,n. \\
        \|\pmb{\mu}_i\| &= \mu \ \forall \ i=1,2,\dots,n.
    \end{align}
\end{subequations}
The overall solution is obtained by first solving for the shape of the tendon-actuated tube and then using its position to solve for the shape of the ball chain. 

\section{Closed-form Kinematics}
\label{sec:inv_kin}
In the prior section, a kinematic model was derived under the assumption that the ball chain does not deflect the proximal tendon-actuated tube. Even with this assumption, solution of the model involves solving a two-point boundary value problem and a constrained energy minimization problem. For purposes of design, workspace analysis and initialization of more complicated models, it is convenient to have a closed-form solution for the kinematics. 

We present such a model by making two additional simplifications. First, as explained in \cite{Rucker2011}, the constant-curvature model for a tendon-actuated tube in which the tendon load is approximated as a point moment at the tip is as accurate as the model of \cite{Rucker2011} for cases of no external load or loads applied in the plane of bending. This condition is satisfied based on our initial no-coupling assumption. 

Second, as noted in \cite{Pittiglio2023b}, the ball chain assumes a nearly linear shape aligned with the external magnetic field for sufficiently high field strengths. Under this assumption, it is possible to derive approximate closed-form forward and inverse kinematic models, as described below.

\subsection{Forward Kinematics}
The following notation is graphically explained in Fig. \ref{fig:hybrid}.
A ball chain robot is assumed to extend, for a length $l_c = n d_c$, as a straight line; here $n$ is the number of balls and $d_c$ their diameter. Given their base position and orientation in global reference frame, $^0\pmb p_b$ and $^0\pmb R_b$, their tip position and orientation is 
\begin{subequations}
\label{eq:chain_kin}
	\begin{align}
		\pmb p &={} ^0\pmb p_b + n d_c \ {}^0\pmb R_b \pmb{\omega}  \label{seq:chain_pos} \\
		\pmb R &= {}^0\pmb R_b \text{exp}\left(\pmb{\omega}_\times \right). \label{seq:chain_rot}
	\end{align}
\end{subequations}
The direction of the chain is a vector of norm 1, $\pmb{\omega} = \hat{\pmb B} = \pmb B/\|\pmb B\|$, and $\text{exp}(\pmb{\omega}_\times)$ the matrix exponential of $\pmb{\omega}_\times$.
We approximate the tendon-actuated continuum robot using constant curvature assumption, 
\begin{subequations}
\label{eq:tube_kin}
	\begin{align}
		\pmb p &={} ^0\pmb p_b + {}^0\pmb R_b \pmb{rot}_{\pmb e_3}(\phi)\pmb{w}   \label{seq:tend_pos} \\
		\pmb R &= {}^0\pmb R_b \pmb{rot}_{\pmb e_3}(\phi)\pmb{rot}_{\pmb e_2}\left(\frac{l_t}{\rho}\right) \label{seq:tend_rot}
	\end{align}
\end{subequations}
with $\pmb{w} = \rho \left(1 - \cos(l_t/\rho) \ 0 \  \sin(l_t/\rho) \right)^T$, $\rho$ and $l_t$ respective radius of curvature and length of the tendon-actuated continuum robot; $\pmb{rot}_{\pmb e_i}(\delta)$ is the rotation of the angle $\delta$ around the axis $\pmb e_i$.

By combining (\ref{eq:chain_kin}) with (\ref{eq:tube_kin}), we obtain the kinematics of the hybrid robot
\begin{subequations}
\label{eq:hyb_kin}
	\begin{align}
		\pmb p &={} ^0\pmb p_b + {}^0\pmb R_b \pmb{rot}_{\pmb e_3}(\phi)\pmb{w} + n d_c \pmb{rot}_{\pmb e_2}(\beta) \pmb \omega  \label{seq:hyb_pos} \\
		\pmb R &= {}^0\pmb R_b \text{exp}\left(\pmb{\omega}_\times \right) \label{seq:hyb_rot}
	\end{align}
\end{subequations}
Notice that, for any orientation of the tip of the tendon-actuated portion, the orientation of the chain's tip is dictated by the magnetic field direction.

\begin{figure*}[t]
    \centering
    \includegraphics[width = \textwidth]{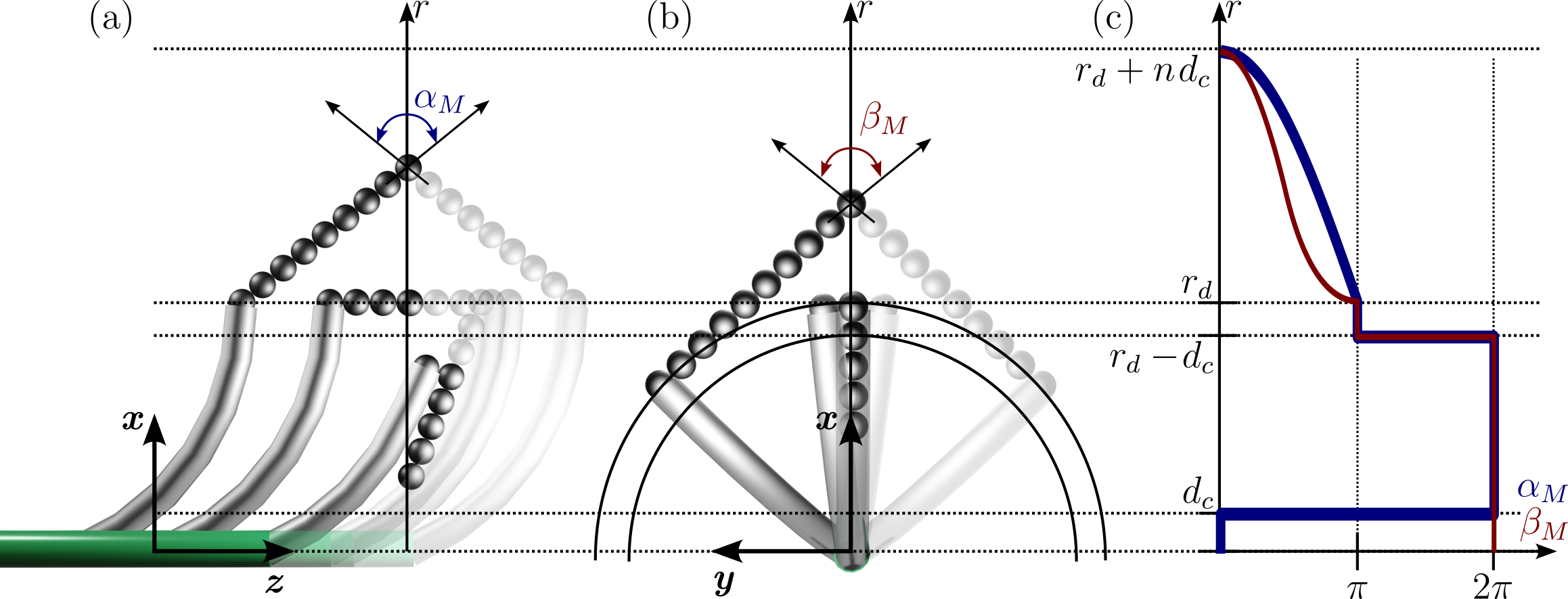}
    \caption{Dexterous workspace. (a) Side view orientation angle range, $\alpha_M$, as function of $x$. (b) Front view orientation angle, $\beta_M$, as function of $x$. (c) $\alpha_M$ and $\beta_M$ as a function of $x$. Dexterous workspace lies in range $d_c \le x \le r_d - d_c$}
    \label{fig:workspace}
\end{figure*}

\subsection{Inverse Kinematics}
The inverse kinematics problem is to solve for the set of inputs $\{F, \phi, \hat{\pmb B}\}$ for which the last ball of the chain approaches a goal position $\pmb p_g$ with orientation $\pmb v_g$ (see Fig. \ref{fig:inv_kin}). The input $F$ refers to the pulling force applied to the tendon, $\phi$ is the angular rotation around the main axis $\pmb z_b$, and $\hat{\pmb B}$ is the field direction. 

First, we guarantee that the objective is feasible, i.e. we check if the point is reachable and the approach angles are within range. The distance from the axis is computed as $r = \sqrt{p_{g_x}^2 + p_{g_y}^2}$, and has to guarantee $r \leq r_d + nd_c$. The goal angle $\alpha_g$ and $\beta_g$ can be found as the twice the angle between $\pmb v_g$ and the respective axes $\pmb e_2$ and $\pmb e_3$: $\alpha_g = 2\arctan(\pmb v_g~\cdot \pmb e_2/||\pmb v_g\times \pmb e_2||)$ and $\beta_g = 2\arctan(\pmb v~\cdot \pmb e_3/||\pmb v_g\times \pmb e_3||)$. If $\alpha_g \leq \alpha_M(r)$ and $\beta_g \leq \beta_M(r)$, the inverse kinematics can be solved. The maximum range of orientations is defined in the next section.

To solve for the tendon-actuated section, we search for the intersection between a line (the ball chain), extending from $\pmb p_g$ in the direction $- \pmb v_g$, and the cylinder of axis $\pmb e_3$ and radius $r_c$:
\begin{equation}
    \left\{
    \begin{array}{lcl}
    \pmb p(s) & = & \pmb p_g - \pmb v_g s  \\
    p_x^2(s) + p_y^2(s) & \leq & r_d \\
    ||\pmb v_g s|| \leq nd_c
    \end{array}
    \right.
\end{equation}
which also satisfies
\begin{equation}
\label{eq:s_select}
\left\{
\begin{array}{ccccc}
      A - B & \leq & s & \leq & A + B \\
     && s & \leq & \frac{n d_c}{||\pmb v_g||}
\end{array}
\right.
\end{equation}
where
\begin{subequations}
\label{eq:constrains}
    \begin{align}
       A & = \frac{-p_{g_x} v_{g_x} + p_{g_y} v_{g_y}}{v_{g_x}^2 + v_{g_y}^2} \\
       B  & =  \frac{\sqrt{-(p_{g_x} v_{g_x} + p_{g_y} v_{g_y})^2 - r_d(v_{g_x}^2 + v_{g_y}^2)}}{v_{g_x}^2 + v_{g_y}^2}
       \end{align}
\end{subequations}

We can select any $s^*$ satisfying (\ref{eq:constrains}) and find the intersection point $\pmb p_i = \pmb p(s^*)$. 

We convert $\pmb p_i$ into cylindrical coordinates to find $\phi = \arctan(p_{i_y}/p_{i_x})$ and, from the direct kinematics in (\ref{eq:tube_kin}), we impose $\rho(1 - \cos(l_t/\rho)) = p_{i_x}$ and obtain $\rho = p_{i_x}/(1 - \cos(k))$, for some $k > 0$. Assuming constant curvature, the moment at the tip of the tube is $\pmb m = EI/\rho \pmb e_2$, and by applying the force $F$ along the surface of the robot, we obtain the equilibrium $F d/2 \pmb e_2 = - EI/\rho \pmb e_2$ and find the tendon force to be
\begin{equation}
    F = -\frac{2EI(1 - \cos(k))}{dp_{i_x}}.
\end{equation}

The constant $k$ is constrained by the inserted length of the robot, depending on the anatomical site, and the amount of tension which can be applied to the tendon. These considerations are fundamental in the design for application-specific catheters.

\section{Dexterous Workspace Characterization}
\label{sec:workspace}
In Fig. \ref{fig:workspace}, we describe the dexterous workspace of the hybrid design, derived from (\ref{eq:hyb_kin}), in the $\pmb{x}-\pmb{z}$ plane in Fig. \ref{fig:workspace}(a) and in the $\pmb{x}-\pmb{y}$ plane in Fig. \ref{fig:workspace}(b). Figure \ref{fig:workspace}(c) depicts the range of approach angles that the hybrid robot can achieve with respect to the radial distance $r$, from the $\pmb{z}$ axis. The achievable robot tip tangent directions are described in terms of the allowable angle ranges $\alpha_M$ about the $\pmb y$ axis in the $\pmb x- \pmb{z}$ plane, and $\beta_M$ about the $\pmb{z}$ axis in the $\pmb x - \pmb y$ plane. 

Full dexterity occurs when $\alpha_M=\beta_M= 2\pi$. As shown in Fig. \ref{fig:workspace}(c), this is true in the range $d_c < r < r_d - d_c$. 
Here, we consider the center of the distal ball to be the robot tip.
The distance $r_d$ corresponds to the radius of curvature of the steerable sheath when its tip is deflected by $\pi/2$. If the maximum deflection of the sheath is less than $\pi/2$ then $r_d$ is given by its maximum radial deflection. 
% from the main axis depends on the capabilities of the tendon-actuated continuum robot. When it is possible to bend it beyond a tip bending $\theta_d = \pi/2$ at the tip, $r_d$ is the radius of curvature for bending of $\pi/2$, i.e. $r_d = l_t/\theta_d$; otherwise $r_d$ is the minimum bending radius. The reason behind this is that, if the bending radius decreases beyond $r_d$, the reachable workspace decreases, without any benefit in dexterity. In fact, every point of the dexterous workspace $d_c \leq r \leq r_d - d_c$ can already be reached from any direction.

In the range $r_d - d_c \leq r \leq r_d$, the ball chain cannot achieve tangent directions pointing downward and so $\alpha_M=\beta_M=\pi$.

For $r > r_d$, the maximum angular ranges $\alpha_M$ and $\beta_M$ can be found using trigonometric approaches. The former, as already derived in \cite{Pittiglio2023b}, is found as twice the angle between the triangle's hypotenuse ($nd_c$) and the side ($r - r_d$): $r - r_d = nd_c \cos(\alpha_M/2)$ thus, 
\begin{equation}
    \alpha_M = 2\arccos\left(\frac{r - r_d}{n d_c}\right), \ r \geq r_d.
\end{equation}
Notice that, in this case, we could assume the triangle is a right triangle. On the contrary, in Fig. \ref{fig:workspace}(b), this assumption is not valid, in general, and we use the law of cosines: $r_d^2 = r^2 + n^2d_c^2 - 2rnd_c\cos(\beta_M/2)$. We obtain the angular range 
\begin{equation}
    \beta_M = 2\arccos\left(-\frac{r_d^2 - r^2 - n^2d_c^2}{2rnd_c}\right), \ r \geq r_d.
\end{equation}

\section{Experimental Validation}
\label{sec:experiments}
\begin{figure}[t]
    \centering
    \includegraphics[width = \columnwidth]{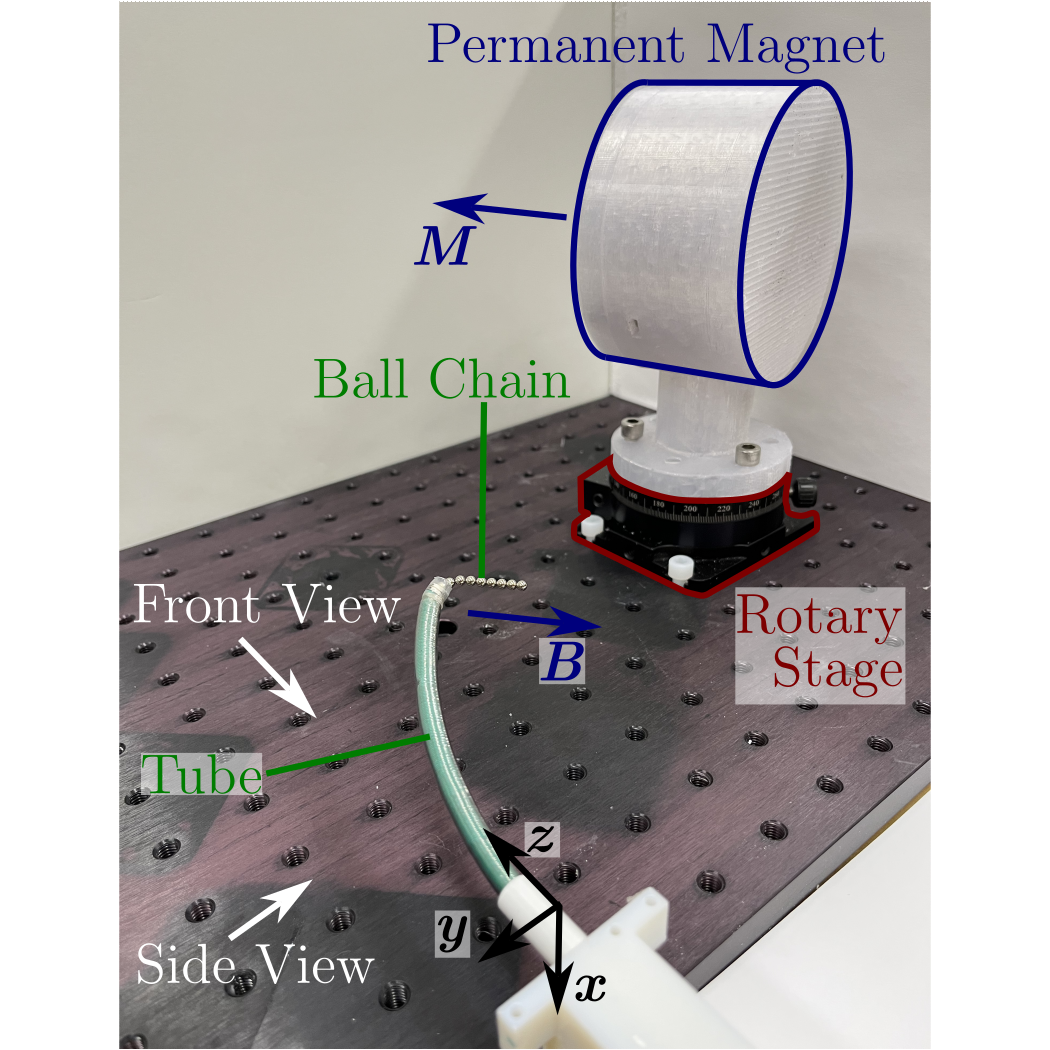}
    \caption{Experiment set up for Front View magnetic field rotations about the $\pmb{x}$ axis.}
    \label{fig:exp_setup}
\end{figure}

We conducted experiments to validate the kinematic model of Section II and to demonstrate the dexterous workspace derived in Section IV.  The data was collected using the setup in Fig. \ref{fig:exp_setup}, consisting of two orthogonal cameras (webcam C920, Logitech, US) placed in the directions of the front and side views as labeled. 

The tendon-actuated tube was 10.16cm (4 inch) long with the tendon positioned upwards (in the $\pmb x$ direction). 
The tube is composed of two layers: an inner layer of PTFE (polytetrafluoroethylene), an outer layer made of PEBAX (a modified polyamide). A braid made of Stainless Steel 304, is placed between them. The tendon runs the entire length between the outer layer and the braid. 

The ball chain was composed of ten N42 magnetic spheres of diameter 3.175mm (1/8inch), mass 0.13g and remanence 1.32T (K\&J Magnetics, USA).
An external N52 cylindrical magnet (76.2mm diameter, 38.1mm long, 1.48T remanence) was mounted on a rotatory stage so that it could be precisely positioned with respect to the robot, and its dipole direction could be precisely controlled.

The tube was loaded with no load (0Kg), medium load (0.4Kg) and max load (1.4Kg), by hanging weights over pulleys at the proximal end of the tendon. The maximum load was selected to achieve a bending angle of 90$^\circ$ at the tube's tip. We repeated the experiments with the same loads while extending ball chains of four and eight balls, and rotating the external magnet to four orientations: $\pm \pmb y, \ \pm \pmb z$. The overall set of data used for validation included 24 different configurations. We also collected data (three repetitions) with the tube without internal magnetic balls and used it to compute the elastic modulus of the tube, found to be approximately 4.10GPa, and a shear modulus of 34.13MPa. The tube was observed to retain a radius of curvature of 56.4mm when the tendon load was removed. Also, the plane of bending differed from the anticipated vertical plane by an angle of $-26.8^\circ$ rotation around the $\pmb z$ axis from the $\pmb x$ direction. These parameters were estimated by tracking the tube's center line using the orthogonal set of webcams.

The tubes' center line was extracted from the images using the \emph{active contour} and \emph{skeletonizing} operations from the Matlab Image Processing Toolbox. 
The position of each ball was found using \emph{imfindcircles}. The data was converted from the image space to the world space by calibrating each camera using the Matlab \emph{camera calibrator} app; the calibration error was found to be of 1.20px for the front camera and 1.46px for the side camera.

\subsection{Kinematic Model Evaluation}

We evaluated the kinematic model of Section II by comparing the experimental data with model predictions. To assess the assumption that the ball chain does not affect tube shape, we computed the error for both tube tip position and ball chain tip position. 

While the tube's centerline was accurately extracted from the images as described above, the tip of the tube was not accurately identified from the images. In contrast, the centers of all extended balls were accurately determined. Therefore, we used the center position of the proximal extended ball as a proxy for tube tip position in order to compute tube modeling error. The center position of the distal ball was used to assess overall error in the combined tube and ball chain model. 

%To validate the assumption of decoupling between magnetic ball chain and tube, we computed the error between the tip of the tube during the experiments and its average position during loading experiments without magnetic balls. For each loading condition, we compute the average position of the tube's tip and compare to the one measured without magnetic interaction (no balls and applied field, average over three repetitions). We found an average error of 6.2mm (max 8.5mm), which is comparable to the size of the tip of the catheter (6mm), thus within the error of the segmentation algorithm. Therefore, we can notice negligible interaction between the tendon-actuated and magnetically-actuated sections of the robot. 

\begin{figure}
    \centering
    \includegraphics[width = \columnwidth]{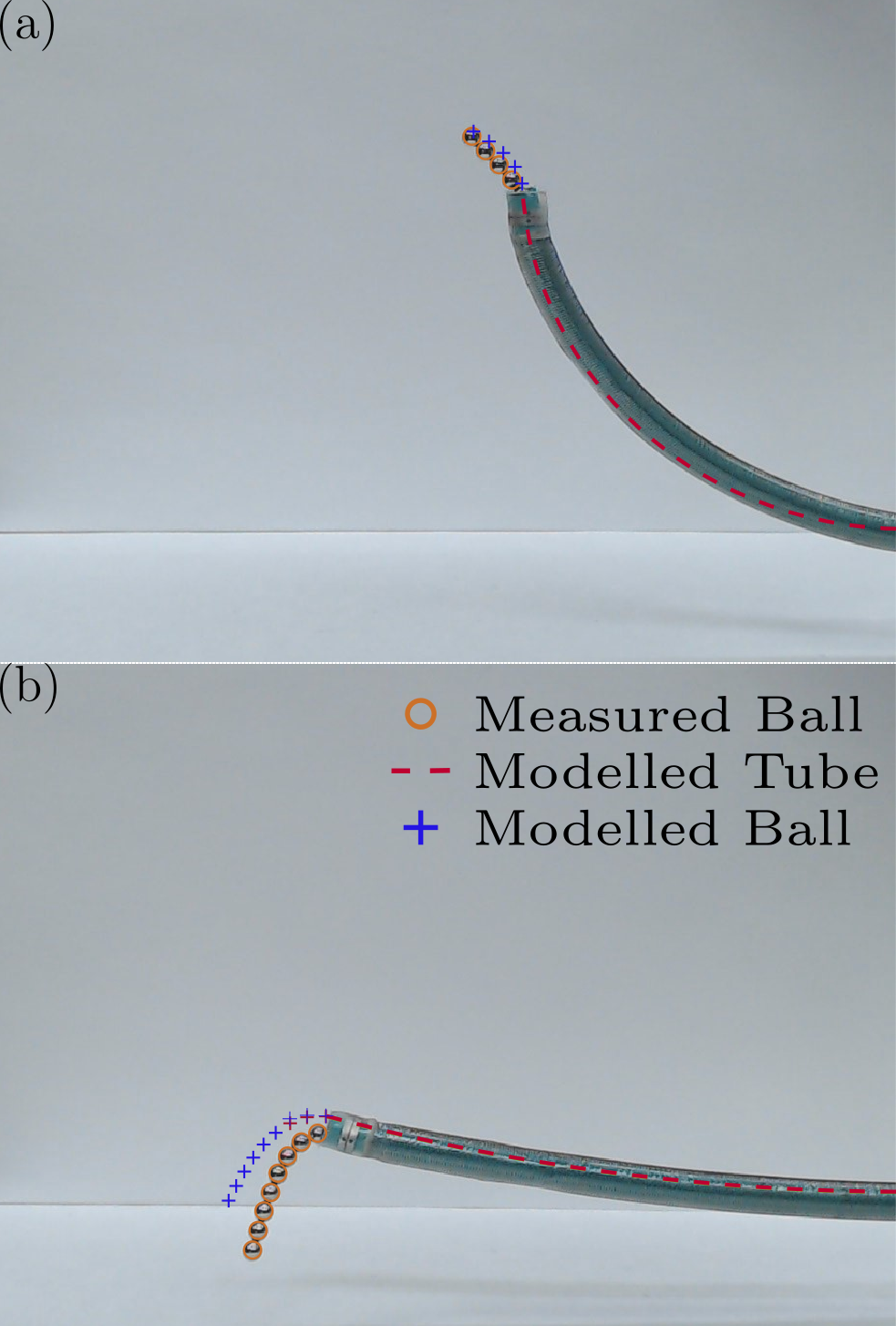}
    \caption{Side view experimental results for configurations with (a) smallest error; (b) largest error.}
    \label{fig:res_pics}
\end{figure}

Table I reports the mean and maximum modeling errors for the tube and total models. We also include the errors as percentages normalized with respect to the tube's (Tube Model) and total robot's (Total Model) length. Mean errors of 1.5-3\% of length have been reported for other types of continuum robots \cite{Grassmann2022}. Given that our model assumes constant curvature tube bending and no tube deflection from the ball chain, a total model mean error of 3\% is quite good. For clinical use, the maximum error is perhaps the more important quantity to consider and a value of 8.7mm for a 132mm long robot could be too large for some applications. (See Fig. \ref{fig:res_pics} for depictions of minimum and maximum error configurations.) 

To improve the accuracy, a coupled model that includes deflection of the tube due the ball chain is likely to substantially reduce the error. This can be inferred by comparing the tube and total model errors in Table I. The mean error in the tube model represents 84\%  of the total model error. In addition, during the experiments, deflection of the tube was observed during motion of the external permanent magnet.

% The lowest is reported in Fig. \ref{fig:res_pics}(a) and the highest in Fig. \ref{fig:res_pics}(b), and we notice that it is mostly related to the inaccuracies in predicting the tube's tip position, in fact the total model error is close to the tube model error. Inaccuracies arise from small (5\% max error) deflections related to the interaction between the chain and the tube, which we did not consider, to simplify the model and reduce its computational intensity.%and its related to the force that the external magnet applies on the chain which opposes to the bending of the tendon actuated tube. As we see in Fig. \ref{fig:res_pics}(a), when the chain direction is tangent to the tube tip we observe minor errors in predicting the tube's shape. 

\begin{table}[]
\centering
\caption{Kinematic Modeling Error. Percentage is tip error normalized by robot length.}
\label{table}
\begin{tabular}{l|c|c|}
\cline{2-3}
                                          & \textbf{Tube Model}                  & \textbf{Total Model}                 \\ \hline
\multicolumn{1}{|l|}{\textbf{Mean Error}} &  3.7mm (3.8\%)                        & 4.4mm (3.0\%)                      \\ \hline
\multicolumn{1}{|l|}{\textbf{Max Error}} & \multicolumn{1}{l|}{4.9mm (5.0\%)} & \multicolumn{1}{l|}{8.7mm (7.0\%)} \\ \hline
\end{tabular}
\end{table}

% Overall, we observe that the proposed hybrid model, combination of Cosserat rod and magnetic energy minimization, is accurate enough (error $\leq$ 7.0\%) in predicting the robot's tip position. We expect that a full model which considers chain-tube full interaction will further reduce the modelling errors and provide more accurate tip tracking capabilities.

\subsection{Dexterous Workspace Demonstration}
To assess the dextrous workspace of the hybrid design as shown in Fig.  \ref{fig:workspace}, we placed a target in the dexterous portion of the workspace located along the main axis ($\pmb{z}$) of the catheter, and displaced 25mm upwards (axis $-\pmb x$). We then positioned the robot such that the tip touched the target from directions ranging over $2\pi$ in the front and side view planes. 

\begin{figure}[t]
    \centering
    \includegraphics[scale = 0.92]{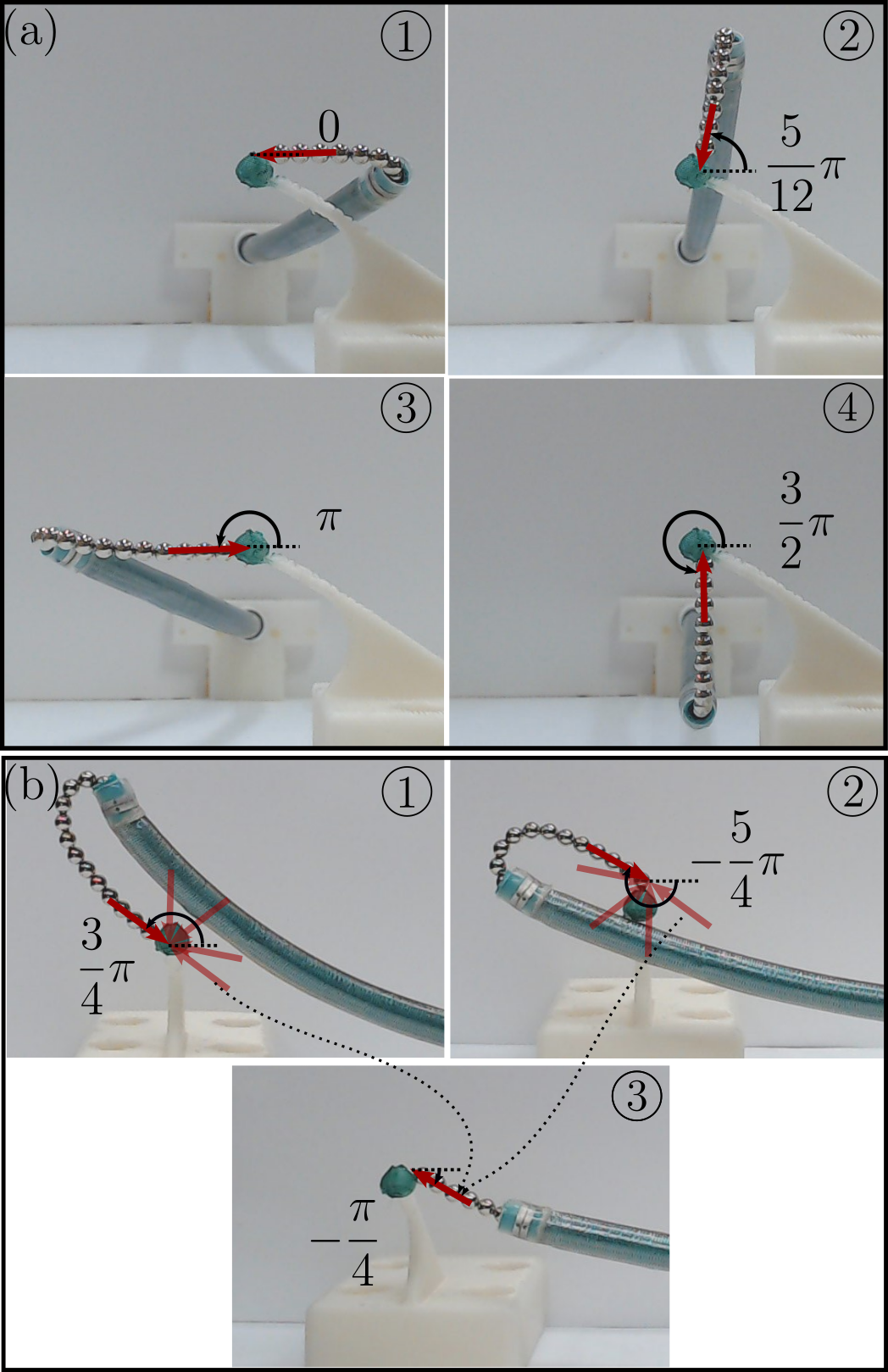}
    \caption{Assessing tip dexterity.
    (a) Front view configurations spaced $\sim\pi/2$ apart illustrating that $\beta_M = 2\pi$. (b) Side view configurations. While $\alpha_M=2\pi$, side view tip approach angles are grouped into ball chain down \protect\numcircled{1} and ball chain up \protect\numcircled{2} configurations depicted here at their common tip tangent boundary. The red arrows indicate the range of tip tangent angles associated with each set. To switch from one set to the other, it is necessary to pass through configuration \protect\numcircled{3}.}
    
    \label{fig:target}
\end{figure}

Four configurations of the front view are shown in Fig. \ref{fig:target}(a) that are spaced roughly $\pi/2$ apart illustrating that $\beta_M = 2\pi$. Note that the hybrid robot can rotate continuously through this angle range. 

In contrast, the side view configurations (Fig. \ref{fig:target}(b)) experience a geometric constraint in that the ball chain cannot pass physically through the curved tube. While $\alpha_M=2\pi$, this results in a tip tangent angle, as shown in Fig. \ref{fig:target}(b) \numcircled{1} and \numcircled{2}, that can be approached from either above or below. The tip tangent cannot, however, move continuously through this value, but instead has to switch from ball chain up to ball chain down or vice versa by passing through the configuration of Fig. \ref{fig:target}(b) \numcircled{3}.

% Specifically in Fig. \ref{fig:target}(a) the catheter bends on the $\pmb x$-$\pmb z$ plane. Figure \ref{fig:target}(b) shows different configurations with catheter bending i
% n the $\pmb x$-$\pmb y$ plane. The maximum angle of about $2\pi$ is reached in Fig. \ref{fig:target}(a), image \numcircled{4} and \ref{fig:target}(b), image \numcircled{4}.
Figure \ref{fig:target} demonstrates that the hybrid continuum robots proposed here do have a dexterous workspace, even if the external magnetic field does not always suffice to fully straighten the ball chain as is assumed by the model of Section IV. This provides the capability for the robot to approach points in its dextrous workspace from all possible directions. 
% In particular, one can observe that the approximation becomes less accurate as the number of balls extended from the sheath increases. However, the range of orientation angles respects the workspace analysis in Section \ref{sec:workspace}.

\section{Conclusions}
In the present paper, we introduced a novel concept for designing continuum robots to produce a dexterous workspace. This hybrid approach combines a tendon-actuated proximal sheath with a distal telescoping magnetic ball chain controlled using an external magnet. 

While a complete mechanics-based model is outlined, the paper develops a simpler model that enables sequential solution of tube shape followed by ball chain shape. The simpler model is tested experimentally and shown to provide normalized mean tip error comparable to other published physics-based continuum robot models. 

For evaluating workspace dexterity, we also describe a further simplified kinematic model that treats the sheath as constant curvature and the ball chain as linear. This results in a closed-form solution for the forward and inverse kinematics. Experiments confirm the validity of our dexterity analysis.

Future work will evaluate feedforward and feedback control of these hybrid designs. This will include a comparison of simplified and coupled mechanics-based models as well as the application of the system to potential clinical tasks.

%\section*{Acknowledgment}

\bibliographystyle{IEEEtran}
\bibliography{bibliography}											

\begin{thebibliography}{1}
\providecommand{\url}[1]{#1}
\csname url@samestyle\endcsname
\providecommand{\newblock}{\relax}
\providecommand{\bibinfo}[2]{#2}
\providecommand{\BIBentrySTDinterwordspacing}{\spaceskip=0pt\relax}
\providecommand{\BIBentryALTinterwordstretchfactor}{4}
\providecommand{\BIBentryALTinterwordspacing}{\spaceskip=\fontdimen2\font plus
\BIBentryALTinterwordstretchfactor\fontdimen3\font minus
  \fontdimen4\font\relax}
\providecommand{\BIBforeignlanguage}[2]{{%
\expandafter\ifx\csname l@#1\endcsname\relax
\typeout{** WARNING: IEEEtran.bst: No hyphenation pattern has been}%
\typeout{** loaded for the language `#1'. Using the pattern for}%
\typeout{** the default language instead.}%
\else
\language=\csname l@#1\endcsname
\fi
#2}}
\providecommand{\BIBdecl}{\relax}
\BIBdecl

\bibitem{Dupont2022}
P.~Dupont, N.~Simaan, H.~Choset, and C.~Rucker, ``{Continuum Robots for Medical
  Interventions},'' \emph{Proceedings of the IEEE}, 2022.

\bibitem{Rucker2011}
D.~C. Rucker and R.~J. Webster, ``{Statics and dynamics of continuum robots
  with general tendon routing and external loading},'' \emph{IEEE Transactions
  on Robotics}, vol.~27, no.~6, pp. 1033--1044, 2011.

\bibitem{Pittiglio2023a}
G.~Pittiglio, M.~Mencattelli, and P.~E. Dupont, ``Magnetic ball chain robots
  for endoluminal interventions,'' in \emph{2023 IEEE International Conference
  on Robotics and Automation (ICRA)}, 2023, pp. 4717--4723.

\bibitem{ODonoghue2013}
K.~O'Donoghue and P.~Cantillon-Murphy, ``Deflection modeling of permanent
  magnet spherical chains in the presence of external magnetic fields,''
  \emph{Journal of Magnetism and Magnetic Materials}, vol. 343, pp. 251--256,
  2013.

\bibitem{Pittiglio2023b}
G.~Pittiglio, M.~Mencattelli, and P.~E. Dupont, ``Closed-form kinematic model
  and workspace characterization for magnetic ball chain robots,'' in
  \emph{2023 International Symposium on Medical Robotics (ISMR)}, 2023, pp.
  1--7.

\bibitem{Grassmann2022}
R.~M. Grassmann, R.~Z. Chen, N.~Liang, and J.~Burgner-Kahrs, ``A dataset and
  benchmark for learning the kinematics of concentric tube continuum robots,''
  in \emph{2022 IEEE/RSJ International Conference on Intelligent Robots and
  Systems (IROS)}, 2022, pp. 9550--9557.

\end{thebibliography}
\end{document}